\documentclass[letterpaper]{article}
\usepackage{aaai20}
\usepackage{times}
\usepackage{helvet}
\usepackage{courier}
\usepackage[hyphens]{url}
\usepackage{graphicx}
\usepackage{graphicx}
\usepackage[utf8]{inputenc} 
\usepackage[T1]{fontenc}    
\usepackage{url}            
\usepackage{booktabs}       
\usepackage{amsfonts}       
\usepackage{nicefrac}       
\usepackage{microtype}      
\usepackage{todonotes}      
\usepackage{amsmath, amssymb, graphicx, braket}
\usepackage[ruled,vlined,linesnumbered]{algorithm2e}

\usepackage{amsthm}
\usepackage{caption}
\usepackage{subcaption}
\usepackage{pdfpages}
\usepackage{arydshln}
\theoremstyle{definition}
\newtheorem{definition}{Definition}
\DeclareMathOperator*{\argmax}{arg\,max}

\usepackage{placeins}

\title{Reasoning on Knowledge Graphs with Debate Dynamics}

\author{Marcel Hildebrandt,\textsuperscript{\rm 1, 2,}\thanks{These authors contributed equally to this work.}
Jorge Andres Quintero Serna,\textsuperscript{\rm 1, 2,$*$}
Yunpu Ma,\textsuperscript{\rm 1,2}
\\ \bf \Large Martin Ringsquandl,\textsuperscript{\rm 1}
Mitchell  Joblin, \textsuperscript{\rm 1} Volker Tresp \textsuperscript{\rm 1, 2}\\
\textsuperscript{\rm 1}Siemens Corporate Technology,
\textsuperscript{\rm 2}Ludwig Maximilian University \\
\{firstname.lastname\}@siemens.com}

\if false
\author{
   Marcel Hildebrandt\thanks{These authors contributed equally to this work.} \\
   Siemens AG, Corporate Technology \\  Ludwig Maximilian University \\
   marcel.hildebrandt@siemens.com
   \And
   Jorge Andres Quintero Serna\textsuperscript{$*$} \\
   Siemens AG, Corporate Technology \\  Ludwig Maximilian University \\
   jorge.quinterno\_serna@siemens.com
   \And
   Yunpu Ma \\
   Siemens AG, Corporate Technology \\  Ludwig Maximilian University \\
   yunpu.ma@siemens.com
   \AND
   Martin Ringsquandl \\
   Siemens AG, Corporate Technology \\
   martin.ringsquandl@siemens.com
   \And
   Mitchell  Joblin \\
   Siemens AG, Corporate Technology \\
   mitchell.joblin@siemens.com
   \And
   Volker Tresp \\
   Siemens AG, Corporate Technology \\  Ludwig Maximilian University \\
   volker.tresp@siemens.com
}
\fi

\begin{document}

\maketitle

\begin{abstract}
We propose a novel method for automatic reasoning on knowledge graphs based on debate dynamics.  The main idea is to frame the task of triple classification as a debate game between two reinforcement learning agents which extract arguments -- paths in the knowledge graph -- with the goal to promote the fact being true (thesis) or the fact being false (antithesis), respectively. Based on these arguments, a binary classifier, called the judge, decides whether the fact is true or false. The two agents can be considered as sparse, adversarial feature generators that present interpretable evidence for either the thesis or the antithesis. In contrast to other black-box methods, the arguments allow users to get an understanding of the decision of the judge. Since the focus of this work is to create an explainable method that maintains a competitive predictive accuracy, we benchmark our method on the triple classification and link prediction task. Thereby, we find that our method outperforms several baselines on the benchmark datasets FB15k-237, WN18RR, and Hetionet. We also conduct a  survey and find that the extracted arguments are informative for users.
\end{abstract}


\section{Introduction}
\label{sec:introduction}

\if false
\begin{itemize}
    \item use pretrained embeddings for all datasets (also hetionet)
    \item stress more that it is not about outperforming other methods, but rather comprehensible and user interaction
    \item justify the experimental setup
    \item figures for each relation
    \item can we do experiments on WN on all relations?
    \item Furthermore, in order to obtain insight into the mechanics of the system, we will manually pass selected arguments to the judge and inspect how they are influencing the judge's decision.
\end{itemize}
\fi

A large variety of information about the real world can be expressed in terms of entities and their relations. Knowledge graphs (KGs) store facts about the world in terms of triples $(s, p, o)$, where $s$ (subject) and $o$ (object) correspond to nodes in the graph and $p$ (predicate) denotes the edge type connecting both. The nodes in the KG represent entities of the real world and predicates describe relations between pairs of entities.

 KGs are useful for various artificial intelligence (AI) tasks in different fields such as named entity disambiguation in natural language processing \cite{han2010structural}, visual relation detection \cite{baier2017improving}, or collaborative filtering \cite{hildebrandt2019recommender}. Examples of large-size KGs include Freebase  \cite{bollacker2008freebase} and YAGO \cite{suchanek2007yago}. In particular, the Google Knowledge Graph \cite{KGblogpost} is a well-known example of a comprehensive KG with more than 18 billion facts, used in search,  question answering, and various NLP tasks. One major issue, however, is that most real-world KGs are incomplete (i.e., true facts are missing) or contain false facts. Machine learning algorithms designed to address this problem try to infer missing triples or detect false facts based on observed connectivity patterns. Moreover, many tasks such as question answering or collaborative filtering can be formulated in terms of predicting new links in a KG (e.g., \cite{lukovnikov2017neural}, \cite{hildebrandt2018configuration}). Most machine learning approaches for reasoning on KGs embed both entities and predicates into low dimensional vector spaces. A score for the plausibility of a triple can then be computed based on these embeddings. Common to most embedding-based methods is their black-box nature since it is hidden to the user what contributed to this score. This lack of transparency constitutes a potential limitation when it comes to deploying KGs in real world settings. Explainability in the  machine learning community has recently gained attention; in many countries laws that require explainable algorithms have been put in place \cite{goodman2017european}. Additionally, in contrast to one-way black-box configurations, comprehensible machine learning methods enable the construction of systems where both machines and users interact and influence each other. 
 
 Most explainable AI approaches can be roughly categorized into two groups: Post-hoc interpretability and integrated transparency \cite{dovsilovic2018explainable}. While post-hoc interpretability aims to explain the outcome of an already trained black-box model (e.g., via layer-wise relevance propagation \cite{montavon2017explaining}), integrated transparency-based methods either employ internal explanation mechanisms or are naturally explainable due to low model complexity (e.g., linear models). Since low complexity and prediction accuracy are often conflicting objectives, there is typically a trade off between performance and explainability. The goal of this work is to design a KG reasoning method with integrated  transparency that does not sacrifice performance while also allowing a human-in-the-loop.

In this paper we introduce R2D2 (\textbf{R}eveal \textbf{R}elations using \textbf{D}ebate \textbf{D}ynamics), a novel method for triple classification based on reinforcement learning. Inspired by the concept outlined in \cite{irving2018ai} to increase AI safety via debates, we model the task of triple classification as a debate between two agents, each presenting arguments either in favor of the thesis (the triple is true) or the antithesis (the triple is false). Based on these arguments, a binary classifier, called the judge, decides whether the fact is true or false. 
As opposed to most methods based on representation learning, the arguments can be displayed to users such that they can trace back the classification of the judge and potentially overrule the decision or request additional arguments. Hence, the integrated transparency mechanism of R2D2 is not based on low complexity components, but rather on the automatic extraction of interpretable features. While deep learning made manual feature engineering to great extends redundant, this advantage came at the cost of producing results that are difficult to interpret. Our work is an attempt to close the circle by employing deep learning techniques to automatically select sparse, interpretable features. The major contributions of this work are as follows.

\begin{itemize}
    \item To the best of our knowledge, R2D2 constitutes the first model based on debate dynamics for reasoning on KGs.
    \item We benchmark R2D2 with respect to triple classification on the datasets FB15k-237 and WN18RR. Our findings show that R2D2 outperforms all baseline methods with respect to the accuracy, the PR AUC, and the ROC AUC, while being more interpretable.
    \item To demonstrate that R2D2 can in principle be employed for KG completion, we also evaluate its link prediction performance on a subset of FB15k-237. To include a real world task, we employ R2D2 on Hetionet for finding gene-disease associations and new target diseases for drugs. R2D2 either outperforms or keeps up with the performance of all baseline methods on both datasets with respect to standard measures such as the MRR, the mean rank, and hits@k, for k = 3, 10.
    \item We conduct a survey where respondents take the role of the judge classifying the truthfulness of statements solely based on the extracted arguments. Based on a majority vote, we find that nine out of ten statements are classified correctly and that for each statement the classification of the respondents agrees with the decision of R2D2's judge. These findings indicate that the arguments of R2D2 are informative and the judge is aligned with human intuition.
\end{itemize}
This paper is organized as follows. We briefly review KGs and the related literature in the next section. Section \ref{sec:our_method} describes the methodology of R2D2. Section \ref{sec:experiments} details an experimental study on the benchmark datasets FB15k-237, WN18RR, and Hetionet. In particular, we compare R2D2 with various methods from the literature and describes the findings of our survey. In Section \ref{sec:discussion} the quality of the arguments and future works are discussed. We  conclude in Section \ref{sec:conclusion}.

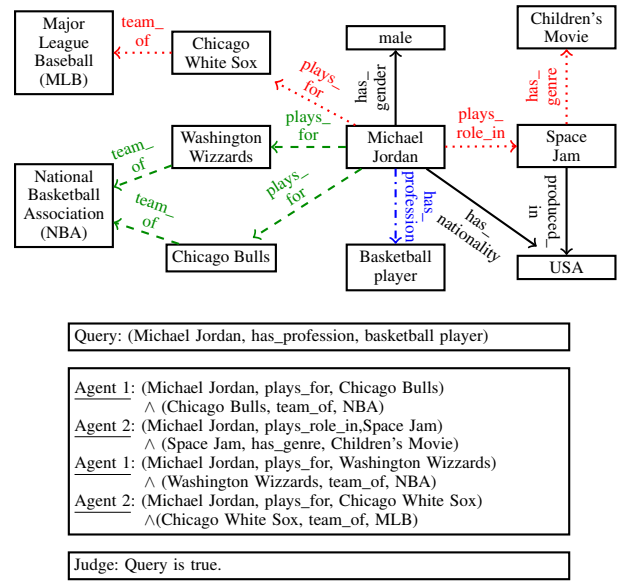
\begin{figure}
\begin{center}
        \begin{tikzpicture}[thick,scale=1,every node/.style={scale=0.65}]
    	\tikzset{state/.style={ draw, minimum width=2cm}}
    	\node[state] (MJ)[align=center] {Michael \\ Jordan};
    	\node[state] (WW) [left=of MJ, align=center]{Washington \\Wizzards};
    	\node[state] (CB) [below=of WW]{Chicago Bulls};
    	\node[state] (NBA) [left=of WW, align=center, yshift=-1.2cm, xshift=0.35cm]{National \\ Basketball \\ Association \\ (NBA)};
    	\node[state] (CWS) [above=of WW, yshift=-0.5cm, align=center]{Chicago \\White Sox};
    	\node[state] (MLB) [left=of CWS, align=center, xshift=0.35cm]{Major \\League \\ Baseball \\ (MLB)};
    	\node[state] (male) [above=of MJ]{male};
        \node[state] (BP) [below=of MJ, align=center]{Basketball \\ player};
    	\node[state] (USA) [right=of BP, xshift=-0.07cm]{USA};
    	\node[state] (SJ) [above=of USA, align=center, yshift=0.25cm]{Space \\Jam};
    	\node[state] (CM) [above=of SJ, align=center]{Children's\\ Movie};

\node[draw,text width=10cm] (Q) [below=of CB, yshift=0.5cm, xshift=2cm] {Query: (Michael Jordan, has\_profession, basketball player) };
\node[draw,text width=10cm, ] (D) [below=of Q, yshift=1.2cm] {

\underline{Agent 1}: (Michael Jordan, plays\_for, Chicago Bulls)  \\ $\quad \quad \quad \; \; \,$ $\wedge$  (Chicago Bulls, team\_of, NBA) \\

\underline{Agent 2}: (Michael Jordan, plays\_role\_in,Space Jam)   \\ $\quad \quad \quad \; \; \,$ $\wedge$ (Space Jam, has\_genre, Children's Movie)

\underline{Agent 1}: (Michael Jordan, plays\_for, Washington Wizzards)\\ $\quad \quad \quad \; \; \,$ $\wedge$ (Washington Wizzards, team\_of, NBA)

\underline{Agent 2}: (Michael Jordan, plays\_for, Chicago White Sox) \\ $\quad \quad \quad \; \; \,$ $\wedge$(Chicago White Sox, team\_of, MLB)
};

\node[draw,text width=10cm] (J) [below=of D, yshift=1.2cm] {Judge: Query is true. };

    	\path[->, shorten >=0.15cm, dashed, black!45!green, align=center] (MJ) edge node[sloped, above] {plays\_\\for} (CB);
    	\path[->, dashed, black!45!green, align=center] (MJ) edge node[sloped, above] {plays\_\\for} (WW);
    	\path[->, dashed, black!45!green, align=center] (CB) edge node[sloped, above] {team\_\\of} (NBA);
    	\path[->, dashed, black!45!green, align=center] (WW) edge node[sloped, above] {team\_\\of} (NBA);
    	\path[->, shorten >=0.2cm, dotted, red, align=center] (MJ) edge node[sloped, above] {plays\_\\for} (CWS);
    	\path[->, dotted, red] (CWS) edge node[sloped, above, align=center] {team\_\\of} (MLB);
    	\path[->, align=center] (MJ) edge node[sloped, above] {has\_\\gender} (male);
    	\path[->, dotted, red, align=center] (MJ) edge node[sloped, above] {plays\_\\role\_in} (SJ);
    	\path[->, shorten >=0.2cm, align=center] (MJ) edge node[sloped, below] {has\_\\nationality} (USA);
    	\path[->, dotted, red, align=center] (SJ) edge node[sloped, above] {has\_\\genre} (CM);
    	\path[->, align=center] (SJ) edge node[sloped, below] {produced\_\\in} (USA);

    	\path[->, align=center, dash dot, blue] (MJ) edge node[sloped, above] {has\_\\profession} (BP);
    	\end{tikzpicture}
\end{center}

	\hspace{1em}
	\label{fig:kg_tensor}
\caption{The agents debate whether Michael Jordan is a professional basketball player. While agent 1 extracts arguments from the KG supporting the thesis that the fact is true (green), agent 2 argues that it is false (red). Based on the arguments the judge decides that  Michael Jordan is a professional basketball player.}
\label{fig:MJ_debate}
\end{figure}


\section{Background and Related Work}
\label{sec:related_work}
\if false
\begin{itemize}
 \item KGs
 \item  EMBEDDING BASED METHODS (TRANSLATIONAL; FACTORIZATION; CONVE)
 \item  PATHS-BASED METHODS (PATH RANK; DEEP WALK; MINERVA)
 \item definition of fact prediction and link prediction
 \end{itemize}
\fi 
 
In this section we provide a brief introduction to KGs in a formal setting and review the most relevant related work. Let $\mathcal{E}$ denote the set of entities and consider the set of binary relations $\mathcal{R}$. A knowledge graph $\mathcal{KG} \subset \mathcal{E} \times \mathcal{R} \times \mathcal{E} $ is a collection of facts stored as triples of the form $(s, p, o)$ – subject, predicate, and object. To indicate whether a  triple is true or false, we consider the binary characteristic function $ \phi : \mathcal{E} \times \mathcal{R} \times \mathcal{E} \rightarrow \{ 0, 1 \}$. For all $(s,p,o) \in \mathcal{KG}$ we assume $\phi(s,p,o) = 1$ (i.e., a KG is a collection of true facts). However, in case a triple is not contained in $\mathcal{KG}$, it does not imply that the corresponding fact is false but rather unknown (open world assumption). Since most KGs that are currently in use are incomplete in the sense that they do not contain all true triples or they actually contain false facts, many canonical machine learning tasks are related to KG reasoning. KG reasoning can be roughly categorized according to the following two tasks: first, inference of missing triples (KG completion or link prediction), and second, predicting the truth value of triples (triple classification). While different formulations of these tasks are typically found in the literature (e.g., the completion task may involve predicting either subject or object entities as well as relations between a pair of entities), we employ the following definitions throughout this work.

\begin{definition}[Triple Classification and KG completion]
Given a  triple $(s, p, o) \in \mathcal{E} \times \mathcal{R} \times \mathcal{E}$, triple classification is concerned with  predicting the truth value $ \phi ( s,p,o )$. KG completion is the task to rank object entities $o \in \mathcal{E}$ by their likelihood to form a true triple together with a given subject-predicate-pair $(s,p) \in \mathcal{E} \times \mathcal{R}$. \footnote{Throughout this work, we assume the existence of inverse relations. That means for any relation $p\in \mathcal{R}$ there exists a relation $p^{-1} \in \mathcal{R}$ such that  $(s, p, o) \in \mathcal{KG}$ if and only if $(o, p^{-1}, s) \in \mathcal{KG}$. Hence, the restriction to rank object entities does not lead to a loss of generality.}  
\end{definition}
Many machine learning methods for KGs can be trained to operate in both settings. For example, a triple classifier of the form $f: \mathcal{E} \times \mathcal{R} \times  \mathcal{E} \rightarrow [0,1]$ with $f(s,p,o) \approx \phi(s,p,o)$, induces a completion method given by $f(s,p, \cdot): \mathcal{E} \rightarrow [0,1]$, where function values for different object entities can be used to produce a ranking. While the architecture of R2D2 is designed for triple classification, we demonstrate that it can in principle also work in the KG completion setting. 
The performance on both tasks is reported in Section \ref{sec:experiments}. 

Representation learning is an effective and popular technique underlying many KG refinement methods. The basic idea is to project both entities and relations into a low dimensional vector space. Then the likelihood of triples is modelled as a functional on the embedding spaces. Popular completion methods based on representation learning include the translational embedding methods TransE \cite{bordes2013translating} and TransR \cite{lin2015learning} as well as the factorization approaches RESCAL \cite{nickel2011three}, DistMult \cite{distmult}, ComplEx \cite{trouillon2016complex}, and SimplE \cite{kazemi2018simple}. 
Path-based reasoning methods follow a different philosophy. For instance, the Path-Ranking Algorithm (PRA) proposed in \cite{lao2011random} uses for inference a combination of weighted random walks through the graph. In \cite{wenhan_emnlp2017} the reinforcement learning based path searching approach called DeepPath was proposed, where an agent picks relational paths between entity pairs. Recently, and more related to our work, the multi-hop reasoning method MINERVA was proposed in \cite{minerva}. The basic idea in that paper is to display the query subject and predicate to the agents and let them perform a policy guided walk to the correct object entity. The paths that MINERVA produces also lead to some degree of explainability. However, we find that only actively mining arguments for the thesis and the antithesis, thus exposing both sides of a debate, allow users to make a well-informed decision.  Mining evidence for both positions can also be considered as adversarial feature generation, making the classifier (judge) robust towards contradictory evidence or corrupted data. 


\section{Our Method}
\label{sec:our_method}

\begin{figure*}
 \begin{center}
    \includegraphics[width=\textwidth]{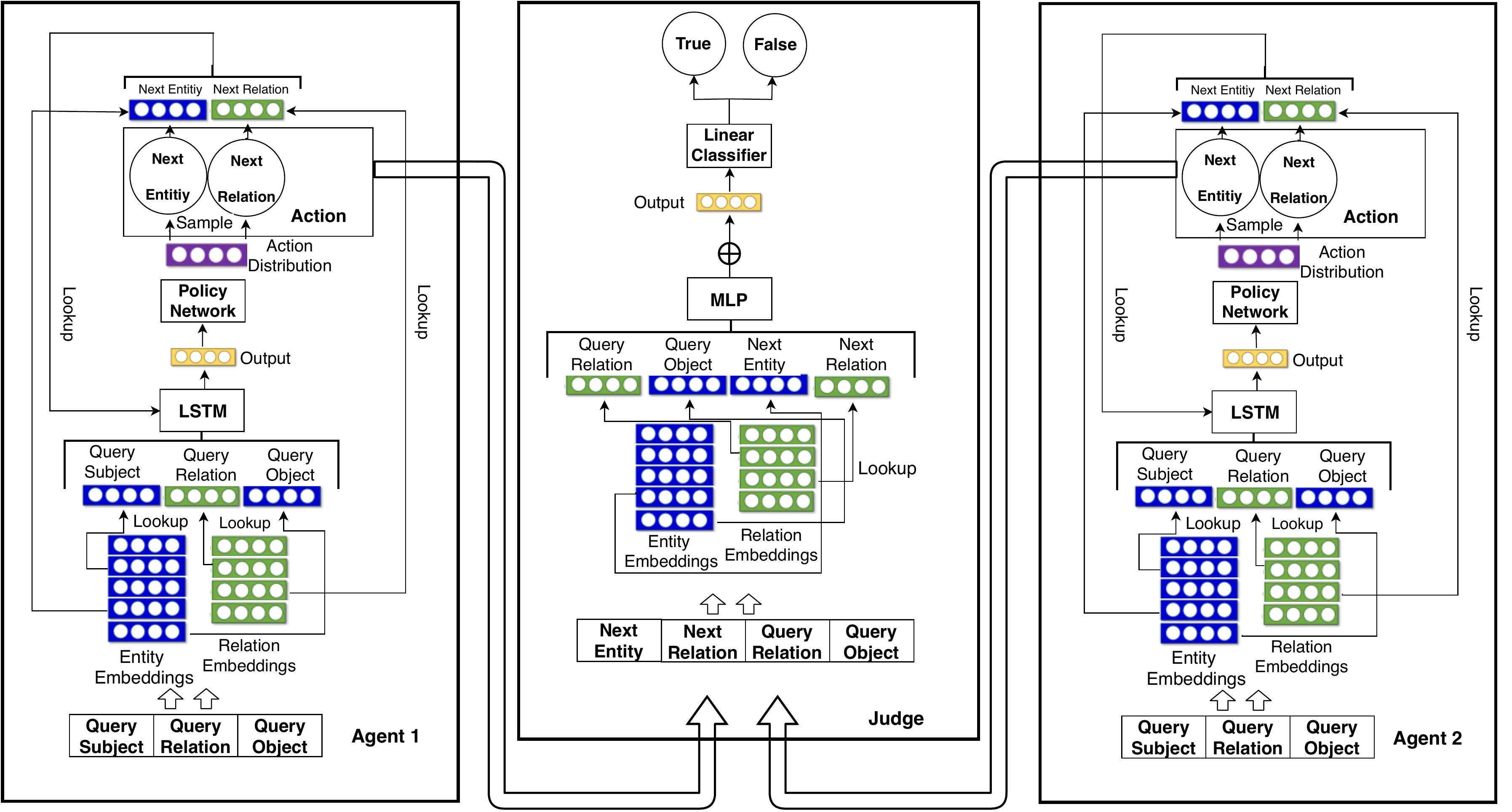}
  \caption{The overall architecture of R2D2; the two agents extract arguments from the KG. Along with the query relation and the query object, these arguments are  processed by the judge who classifies whether the query is true or false. 
  }
  \label{fig:architecture}   
 \end{center}
\end{figure*}


We formulate the task of triple classification in terms of a debate between two opposing agents. Thereby, a query triple corresponds to the statement on which the debate is centered. The agents proceed by mining paths on the KG that serve as evidence for the thesis or the antithesis. More precisely, they traverse the graph sequentially and select the next hop based on a policy that takes past transitions and the query triple into account. The transitions are added to the current path, extending the argument. 
All paths are processed by a binary classifier called the judge that attempts to distinguish between true and false triples based on the arguments provided by the agents. Figure \ref{fig:MJ_debate} shows an exemplary debate. The main steps of a debate can be summarized as follows:
\begin{enumerate}
    \item A query triple around which the debate is centered is presented to both agents.
    \item The two agents take turns extracting paths from the KG that serve as arguments for the thesis and the antithesis.
    \item The judge processes the arguments along with the query triple and estimates the truth value of the query triple.
\end{enumerate}
While the parameters of the judge are fitted in a supervised fashion, both agents are trained to navigate through the graph using reinforcement learning. Generalizing the formal framework presented in \cite{minerva}, the agents' learning tasks are modelled via the fixed horizon decision processes outlined below.

\paragraph{States}
The fully observable state space $\mathcal{S}$ for each agent is given by $\mathcal{E}^2 \times \mathcal{R} \times \mathcal{E}$. Intuitively, we want the state to encode the location of exploration $e^{(i)}_t$ (i.e., the current location) of agent $i \in \{1,2 \}$ at time $t$ and the query triple $q = (s_q,p_q,o_q)$. Thus, a state $S^{(i)}_t \in \mathcal{S}$ for time $t \in \mathbb{N}$ is represented by $S^{(i)}_t = \left(e^{(i)}_t, q\right)$.

\paragraph{Actions} \sloppy
 The set of possible actions for agent $i$ from a state $S^{(i)}_t = \left(e^{(i)}_t, q\right)$ is denoted by $\mathcal{A}_{S^{(i)}_t}$. It consists of all outgoing edges from the node $e^{(i)}_t$ and the corresponding target nodes. More formally, $\mathcal{A}_{S^{(i)}_t} = \left\{(r,e) \in \mathcal{R} \times \mathcal{E} :  S^{(i)}_t = \left(e^{(i)}_t, q\right) \land \left(e^{(i)}_t,r,e\right) \in \mathcal{KG}\right\}\, .$ Moreover, we denote with $A_t^{(i)} \in \mathcal{A}_{S^{(i)}_t}$ the action that agent $i$ performed at time $t$. We include self-loops for each node such that the agent can stay at the current node.

\paragraph{Environments}
The environments evolve deterministically by updating the state according to the agents' actions (i.e., by changing the agents' locations), whereby the query fact remains the same. Formally, the transition function of agent $i$ at time $t$ is given by $\delta_t^{(i)}({S^{(i)}_t},A_t^{(i)}) := \left(e^{(i)}_{t+1}, q\right)$ with $S^{(i)}_t = \left(e^{(i)}_{t}, q\right)$ and $A_t^{(i)} = \left(r, e^{(i)}_{t+1}\right)$.

\paragraph{Policies}
We denote the history of agent $i$ up to time $t$ with the tuple $H^{(i)}_t = \left(H^{(i)}_{t-1}, A^{(i)}_{t-1}\right)$ for $t \geq 1$ and $H^{(i)}_0 = (s_q,p_q,o_q)$ along with $A^{(i)}_0 = \emptyset$ for $t = 0$. The agents encode their histories via LSTMs \cite{hochreiter1997long}
\begin{equation}
\label{eq:lstm_agent}
        \boldsymbol{h}^{(i)}_t = \text{LSTM}^{(i)}\left(\left[\boldsymbol{a}^{(i)}_{t-1}, \boldsymbol{q}^{(i)}\right]\right) \, ,
\end{equation}
where $\boldsymbol{a}^{(i)}_{t-1} = \left[\boldsymbol{r}^{(i)}_{t-1},\boldsymbol{e}^{(i)}_{t}\right] \in \mathbb{R}^{2d}$ corresponds to the vector space embedding of the previous action (or the zero vector for at time $t=0$) with $\boldsymbol{r}^{(i)}_{t-1}$ and $\boldsymbol{e}^{(i)}_{t}$ denoting the embeddings of the relation and the target entity into $\mathbb{R}^{d}$, respectively. Moreover, $\boldsymbol{q}^{(i)} = \left[ \boldsymbol{e}_s^{(i)},\boldsymbol{r}_p^{(i)},\boldsymbol{e}_o^{(i)}\right] \in \mathbb{R}^{3d}$ encodes the query triple for agent $i$. Both entity and relation embeddings are specific for each agent and learned in the debate process during training. Note that expanding the state space definitions with the histories leads to a Markov decision processes.

The history-dependent action distribution of each agent is given by
\begin{equation}
\label{eq:policy_agent}
\boldsymbol{d}_t^{(i)} = \text{softmax}\left(\boldsymbol{A}_t^{(i)} \left(\boldsymbol{W}_2^{(i)}\text{ReLU}\left(\boldsymbol{W}_1^{(i)} \boldsymbol{h}^{(i)}_t\right)\right)\right) \, ,
\end{equation}
where the rows of $\boldsymbol{A}_t^{(i)} \in \mathbb{R}^{\vert \mathcal{A}_{S^{(i)}_t} \vert \times d}$ contain latent representations of all admissible actions from $S_t^{(i)}$. The action $A_t^{(i)} = (r,e) \in \mathcal{A}_{S^{(i)}_t}$ is drawn according to
\begin{equation}
\label{eq:sampling_agent}
    A_t^{(i)} \sim \text{Categorical}\left(\boldsymbol{d}_t^{(i)}\right)\, .
\end{equation}
 Equations \eqref{eq:lstm_agent} and \eqref{eq:policy_agent} define a mapping from the space of histories to the space of distribution over all admissible actions, thus inducing a policy $\pi_{\theta^{(i)}}$, where $\theta^{(i)}$ denotes the set of all trainable parameters in Equations \eqref{eq:lstm_agent} and \eqref{eq:policy_agent}.


\paragraph{Debate Dynamics}
In a first step, the query triple $q = (s_q,p_q,o_q)$ with truth value $\phi(q) \in \{0,1\}$  is presented to both agents. Agent 1 argues that the fact is true, while agent 2 argues that it is false. Similar to most formal debates, we consider a fixed number of rounds $N \in \mathbb{N}$. In every round $n = 1,2, \dots, N$, the agents start graph traversals with fixed length $T \in \mathbb{N}$ from the subject node of the query $s_q$. The judge observes the paths of the agents and predicts the truth value of the triple. Agent 1 starts the game generating a sequence of length $T$ consisting of states and actions according to Equations (\ref{eq:lstm_agent} - \ref{eq:sampling_agent}). Then agent 2 proceeds by producing a similar sequence starting from $s_q$. Algorithm  \ref{alg:debate} contains a pseudocode of R2D2 at inference time.

\begin{algorithm}[t]
\SetAlgoLined
\DontPrintSemicolon
\SetKwInOut{Query}{input}
\SetKwInOut{Judge}{input}
\SetKwInOut{AgentY}{input}
\SetKwInOut{AgentN}{input}
\SetKwInOut{Output}{output}
\Query{Triple query $q=(s_q,p_q,o_q)$}
\Output{Classification score $t_\tau \in [0,1]$ of the judge along with the list of arguments $\tau$}
 $\tau \leftarrow [ \; ]$$\;$\tcp{Initialize the list of arguments with an empty list}
 \tcp{Loop over the debate rounds}
 \For{$n = 1$ to N}{ 
 \tcp{Loop over the two agents}
    \For{$i = 1$ to 2}{
     $e_1^{(i)} \leftarrow s_q$$\;$\tcp{Initialize the position of the agent}
     $\tau_n^{(i)} \leftarrow [ \; ]$$\;$\tcp{Initialize the argument with an empty list}
     \tcp{Loop over the path}
     \For{$t = 1$ to T }{
        Sample a transition $(r,e)$ from $e_t^{(i)}$ according to $\pi_{\theta^{(i)}}$\tcp{See Equations (1-3)}
        $\tau_n^{(i)}\text{.append}(r,e)\;$ \tcp{Extend the argument}
        $e_{t+1}^{(i)} \leftarrow e\;$ \tcp{Update the position of the agent}
     }
     $\tau\text{.append}(\tau_n^{(i)})\;$ \tcp{Extend the list of all argument}}
 }
 Process $\tau$ via the judge and retrieve the classification score $t_\tau \;$\tcp{See Equation (6-8)}
 \KwRet{$t_\tau$ and $\tau$}
 \caption{R2D2 Inference}
 \label{alg:debate}
\end{algorithm}

To ease the notation we have enumerated all actions consecutively and dropped the superscripts that indicate which agent performs the action. Then the sequence corresponding to the $n$-th argument of agent $i$ is given by
\begin{equation}
    \tau^{(i)}_n := \left( A_{ \tilde{n}(i,T) +1}, A_{ \tilde{n}(i,T) +2}, \dots, A_{ \tilde{n}(i,T) +T}  \right) \, ,
\end{equation}
where we used the reindexing $\tilde{n}(i,T):= (2(n-1)+i-1)T$. 
The sequence of all arguments  is denoted by
\begin{equation}
\label{eq:seq_actions}
    \tau :=  \left( \tau^{(1)}_1, \tau^{(2)}_1, \tau^{(1)}_2, \tau^{(2)}_2, \dots, \tau^{(1)}_N, \tau^{(2)}_N \right) \,.
\end{equation}

\paragraph{The Judge}
The role of the judge in R2D2 is twofold: First, the judge is a binary classifier that tries to distinguish between true and false facts. Second, the judge also evaluates the quality of the arguments extracted by the agents and assigns rewards to them. Thus, the judge also acts as a critic, teaching the agents to produce meaningful arguments. The judge processes each argument together with the query individually by a feed forward neural network $f : \mathbb{R}^{2(T+1)d}\rightarrow \mathbb{R}^{d}$, sums the output for each argument up and processes the resulting sum by a binary classifier. More concretely, after processing each argument individually, the judge produces a representation according to
\begin{equation}
\label{eq:NN_Judge}
    \boldsymbol{y}_n^{(i)} = f\left( \left[\boldsymbol{\tau}^{(i)}_n\, ,\boldsymbol{q}^{J} \right] \right)
\end{equation}
with
\begin{equation}
\label{eq:argument_embedding_judge}
    \boldsymbol{\tau}^{(i)}_n := \left[\boldsymbol{a}^{J}_{ \tilde{n}(i,T) +1}, \boldsymbol{a}^{J}_{ \tilde{n}(i,T) +2}, \dots,  \boldsymbol{a}^{J}_{\tilde{n}(i,T) +T}\right]
\end{equation}
where $\boldsymbol{a}^{J}_{t} = \left[\boldsymbol{r}^{J}_t,\boldsymbol{e}^{J}_t\right] \in \mathbb{R}^{2d}$ denotes the judge's embedding for the action $A_t$ and $\boldsymbol{q}^{J} =\left[\boldsymbol{r}_p^{J},\boldsymbol{e}_o^{J}\right] \in \mathbb{R}^{2d}$ encodes the query predicate and the query object. Note that the query subject is not revealed to the judge because we want the judge to base its decisions solely on the agents' actions rather than on the embedding of the query subject. After processing all arguments in $\tau$, the debate is terminated and the judge scores the query triple $q$ with $t_\tau \in (0,1)$  according to
\begin{equation}
    \label{eq:classifier}
    t_\tau = \sigma\left(\boldsymbol{w}^\intercal \text{ReLU}\left(\boldsymbol{W} \sum_{i = 1}^2 \sum_{n = 1}^N \boldsymbol{y}^{(i)}_n\right)\right) \,,
\end{equation}
where $\boldsymbol{W} \in \mathbb{R}^{d \times d}$ and $\boldsymbol{w} \in \mathbb{R}^{d}$ denote the trainable parameters of the classifier and $\sigma(\cdot)$ denotes the sigmoid activation function. We also experimented with more complex architectures where the judge processes each argument in $\tau$ via a recurrent neural network. However, we found that both the classification performance and the quality of the arguments suffered.

The objective function of the judge for a single query $q$ is given by the cross-entropy loss
\begin{equation}
    \label{eq:loss_judge_query}
    \mathcal{L}_q = \phi(q)\log t_\tau + \left(1-\phi(q)\right)(1-\log t_\tau) \,.
\end{equation}
Hence, during training, we aim to minimize the overall loss given by
\begin{equation}
    \label{eq:loss_judge}
    \mathcal{L} = \frac{1}{\vert \mathcal T \vert}\sum_{q \in \mathcal{T}} \mathcal{L}_q \,,
\end{equation}
    where $\mathcal{T}$ denotes the set of training triples. To prevent overfitting, an additional $L_2$-penalization term with strength $\lambda \in \mathbb{R}_{\geq 0}$ on the parameters of the judge is added to Equation \eqref{eq:loss_judge}. 

An overview of the overall architecture of R2D2 is depicted in Figure \ref{fig:architecture}. 

\paragraph{Rewards}

In order to generate feedback for the agents, the judge also processes each argument $ \tau^{(i)}_n$ individually and produces a score according to 
\begin{equation}
\label{eq:judge_rewards}
    t^{(i)}_n = \boldsymbol{w}^\intercal \text{ReLU}\left(\boldsymbol{W} f\left( \left[ \boldsymbol{\tau}^{(i)}_n, \boldsymbol{q}^{J} \right] \right) \right)\, ,
\end{equation}
where both the neural network $f$ as well as the linear weights vector $\boldsymbol{w}$ correspond to the definitions given in the previous paragraph. Thus, $t^{(i)}_n$ corresponds to the classification score of $q$ solely based  on the $n$-th argument of agent $i$. Since agent 1 argues for the thesis and agent 2 for the antithesis, the rewards are given by
\begin{equation}
    R^{(i)}_n = \begin{cases}
 {t}^{(i)}_n &\text{if } i = 1 \\
-{t}^{(i)}_n &\text{otherwise.}
\end{cases}
\end{equation}
Intuitively speaking, this means that the agents receive high rewards whenever they extract an argument that is considered by judge as strong evidence for their position
\paragraph{Reward Maximization and Training Scheme}
We employ REINFORCE \cite{williams1992simple} to maximize the expected cumulative reward  of the agents given by
\begin{equation}
    G^{(i)} := \sum_{n = 1}^N  R^{(i)}_n \, .
\end{equation}
Thus, the agents' maximization problems are given by
\begin{equation}
\label{eq:objective_agent}
    \argmax_{\theta^{(i)}} \mathbb{E}_{q \sim \mathcal{KG}_+}\mathbb{E}_{\tau^{(i)}_1, \tau^{(i)}_2, \dots, \tau^{(i)}_N \sim \pi_{\theta^{(i)}}}\left[G^{(i)} \left\vert\vphantom{\frac{1}{1}}\right. q \right] \, ,
\end{equation}
where $\mathcal{KG}_+$ is the set of training triples that contain in addition to observed triples in $\mathcal{KG}$ also unobserved triples. The rationale is as follows: As KGs only contain true facts, sampling queries from $\mathcal{KG}$ would create a dataset without negative labels. Therefore it is common to create corrupted triples that are constructed from correct triples $(s,p,o)$ by replacing the object with an entity $\tilde o$ to create a false triple $(s,p,\tilde o) \notin \mathcal{KG}$ (see \cite{bordes2013translating}). Rather than creating any kind of corrupt triples, we generate a set of plausible but false triples. More concretely, for each $(s,p,o) \in \mathcal{KG}$ we generate one triple $(s,p,\tilde o) \notin \mathcal{KG}$ with the constraint that $\tilde o$ appears in the database as the object with respect to the relation $p$. More formally, we denote the set of corrupted triples with 
$ \mathcal{KG}_C  := \left\{   (s,p,\tilde o) \, \vert \,  (s,p, \tilde o) \notin \mathcal{KG}, \exists \tilde s: (\tilde s,p,\tilde o) \in \mathcal{KG}  \right\} \, .
$ Then the set of training triples $\mathcal{T}$ is contained in $\mathcal{KG}_+  := \mathcal{KG} \, \cup \, \mathcal{KG}_C$.
The underlying rationale for working with plausible but false facts is that we do not waste resources on triples that break implicit type-constrains. Since this heuristic only needs to be computed once and filters out triples that could easily be discarded by a type-checker, we can focus on the prediction of facts that present more of a challenge.

During training the first expectation in Equation \eqref{eq:objective_agent} is substituted with the empirical average over the training set. The second expectation is approximated by the empirical average over multiple rollouts. We also employ a moving average baseline to reduce the variance. Further, we use entropy regularization with parameter $\beta\in \mathbb{R}_{\geq 0}$ to enforce exploration.

In order to address the problem that the agents require a trained judge to obtain meaningful reward signals, we freeze the weights of the agents for the first episodes of the training. The rationale is that training the judge does not necessarily rely on the agents being perfectly aligned with their actual goals. For example, even if the agents do not extract arguments that correspond to their position, they can still provide useful features that the judge learns to exploit. After the initial training phase, where we only fit the parameters of the judge, we employ an alternating training scheme where we either train the judge or the agents.


\section{Experiments}
\label{sec:experiments}
\if false
\begin{itemize}
 \item  EXPERIMENTAL SETUP + table for the data + mention the filtered setting
 \item  RESULTS
 \item QUALITATIVE RESULTS
 \end{itemize}
\fi

\begin{table}
\center
\resizebox{0.8\columnwidth}{!}{
\begin{tabular}{ c  c  c  c}
 \hline
 Dataset & Entities & Relations & Triples \\
 \hline \hline
FB15k-237 & 14,541 & 237 & 310,116 \\
 WN18RR & 40,943 & 11 & 93,003 \\
 Hetionet & 47,031 & 24 & 2,250,197 
 \vspace{0.2cm}
\end{tabular}
}
\caption{Statistics of the datasets used in the experiments.}
\label{tbl:datasets}
\end{table}

\begin{table*}
\center
\resizebox{0.6\textwidth}{!}{
\begin{tabular}{ c c c c |c c c}
\hline
 Dataset & \multicolumn{3}{c|}{FB15k-237} & \multicolumn{3}{c}{WN18RR} \\ 
 \hline
 Method & Acc & PR AUC & ROC AUC & Acc & PR AUC & ROC AUC \\
 \hline \hline
 DistMult & 0.739 & 0.78 & 0.803 & \bf{0.804} & \bf{0.901} & \bf{0.872} \\
 ComplEx & 0.738 & 0.789 & 0.796 & 0.802 & 0.887 & 0.860 \\
 TransE & 0.673 & 0.727 & 0.736 & 0.69 & 0.794 & 0.732  \\
 TransR & 0.612 & 0.655 & 0.651 & 0.721 & 0.724 & 0.792 \\
 SimplE & 0.703 & 0.733 & 0.756 & 0.722 & 0.812 & 0.742 \\
 R2D2 & \bf{0.751} & \bf{0.86} & \bf{0.848} & 0.726 & 0.821 & 0.808 \\
 \hline
 $\text{R2D2}_+$ & \bf{0.764} & \bf{0.865} & \bf{0.857} & \bf{0.804} & \bf{0.909}& \bf{0.893}
\end{tabular}
}
\caption{The performance on the triple classification task.}
\label{tbl:classification_results}
\end{table*}

\paragraph{Datasets} We measure the performance of R2D2 with respect to the triple classification and the KG completion task on the benchmark datasets FB15k-237 \cite{toutanova2015representing} and WN18RR \cite{dettmers2018convolutional}. To test R2D2 on a real world task we also consider Hetionet \cite{himmelstein2015heterogeneous}, a large scale, heterogeneous graph encoding information about chemical compounds, diseases, genes, and molecular functions. We employ R2D2 for detecting gene-disease associations and finding new target diseases for drugs, two tasks of high practical relevance in the biomedical domain (see \cite{himmelstein2015heterogeneous}). The statistics of all datasets are given in Table \ref{tbl:datasets}.

\paragraph{Metrics and Evaluation Scheme} 
As outlined in Section \ref{sec:related_work}, triple classification aims to decide whether a query triple $(s_q,p_q,o_q)$ is true or false. Hence, it is a binary classification task. For each method we set a threshold $\delta$ obtained by maximizing the accuracies on the validation set. That means, for a given query triple $(s_q,p_q,o_q)$, if its score (e.g., given by Equation \eqref{eq:classifier} for R2D2) is larger than $\delta$, the triple will be classified as true, otherwise as false. Since most KGs do not contain facts that are labeled as false, one needs to sample negative triples. We have generated a set of negative triples: For each observed triple in the validation and test set we create a false but plausible fact (see Section \ref{sec:our_method}).\footnote{The datasets along with the code of R2D2 are available at \url{https://github.com/m-hildebrandt/R2D2}.} We report the accuracy, the PR AUC, and ROC AUC for all methods. Since R2D2 is a stochastic classifier, we can produce multiple rollouts of the same query at inference time and average the resulting classification scores to lower the variance.

Even though the purpose of R2D2 is triple classification, one can turn it into a KG completion method as follows: We consider a range of object entities each producing a different classification score $t_\tau$ given by Equation \eqref{eq:classifier}. Since $t_\tau$ can be interpreted as a measure for the plausibility of a triple, we use the classification scores to produce a ranking. More concretely, we rank each correct triple in the test set against all plausible but false triples (see Section \ref{sec:our_method}). Since this procedure is computational expensive during training (one needs to run multiple debates per training triple to produce a ranking), we select the following relations for training and testing purposes: For FB15k-237 we follow \cite{socher2013reasoning} and consider the relations `profession', `nationality', `ethnicity', and `religion'. Following \cite{himmelstein2015heterogeneous} and \cite{himmelstein2017systematic}, the relations `gene\_associated\_with\_disease' and `compound\_treats\_disease'  are considered for Hetionet. 
We report the mean rank of the correct entity, the mean reciprocal rank (MRR), as well as Hits@k for $k = 1,3,10$ - the percentage of test triples where the correct entity is ranked in the top k. 

In order to find the most suitable set of hyperparameters for all considered methods, we perform cross-validations. Thereby the canonical splits of the datasets into a training, validation, and test set are used. In particular, we ensured that triples that are assigned to the validation or test set (and their respective inverse relations) are not included in the KG during training. The results on the test set of all methods are reported based on the hyperparameters that showed the best performance (based on the highest accuracy for triple classification and the the highest MRR for link prediction) on the validation set. We considered the following hyperparameter ranges for R2D2: The number of latent dimensions $d$ for the embeddings is chosen from the range $\{32, 64 , 128\}$. The number of LSTM layers for the agents is chosen from $\{1,2,3\}$. The the number of layers in the MLP for the judge is tuned in the range $\{1,2,3,4,5\}$. $\beta$ was chosen from $\{0.02, 0.05, 0.1\}$. The length of each argument $T$ was tuned in the range $\{1,2,3\}$ and the number of debate rounds $N$ was set to $3$. Moreover, the $L_2$-regularization strength $\lambda$ is set to 0.02. Furthermore, the number of rollouts during training is given by 20 and 50 (triple classification) or 100 (KG completion) at test time. The loss of the judge and the rewards of the agents were optimized using Adam with learning rate given $10^{-4}$. The best hyperparameter are reported in the supplementary material.


All experiments were conducted on a machine with 48 CPU cores and 96 GB RAM. Training R2D2 on either dataset takes at most 4 hour. Testing takes about 1-2 hours depending on the dataset.



\subsection{Results}
\paragraph{Triple Classification}
 We compare the performance of R2D2 on the triple classifications task against  DistMult, ComplEx, TransE, TransR, and SimplE. The results are displayed in Table \ref{tbl:classification_results}. On FB15k-237, R2D2 outperforms all baselines with respect to the accuracy, the PR AUC, and the ROC AUC. However, on WN18RR the performance of R2D2 is dominated by the factorization methods ComplEx and DistMult by a significant margin. We conjecture that this is due to the sparsity in the dataset. As a remedy we employ pretrained embeddings from TransE that are fixed during training.\footnote{We choose TransE embeddings due to the simple functional relations between entities. These can be easily exploited by R2D2.}
 We denote the resulting method with $\text{R2D2}_+$ and find that it outperforms all other methods with respect to the PR AUC and ROC AUC on WN18RR. We also test $\text{R2D2}_+$ on FB15k-237 and find that it improves the results of R2D2 by only a small margin. This is expected since FB15k-237 is not as sparse as WN18RR.
\paragraph{KG completion} Next to the baselines used for triple classification we also employ the path based link prediction method MINERVA. Note that it is not possible to compute a fair mean rank for MINERVA, since it does not produce a complete ranking of all candidate objects.
\begin{table*}
\center
\resizebox{0.9\textwidth}{!}{
\begin{tabular}{ c  c  c  c  c  c | c  c  c  c   c}
 \hline
 Dataset & \multicolumn{5}{c|}{FB15k-237 (subset)} & \multicolumn{5}{c}{Hetionet (subset)} \\ 
 \hline
 Metrics &  MRR & Mean Rank & Hits@1 & Hits@3 & Hits@10  & MRR & Mean Rank & Hits@1 & Hits@3 & Hits@10\\
 \hline
 \hline
 DistMult & 0.502 & 8.607 & 0.363 & 0.572 & 0.779  & 0.134 & 31.190 &  0.054 & 0.121 & 0.291\\
 ComplEx & 0.521 & 7.477 & 0.383 & 0.587 & 0.806 & 0.148 & 31.439 &  0.061 & 0.141 & 0.325 \\
 TransE & 0.473 & 10.2112 & 0.345 & 0.522 & 0.745 &  0.110 & 35.559 & 0.033 &  0.097 & 0.267 \\
 TransR & 0.543 & 8.737 & 0.391 & 0.635 & 0.816 & 0.144 & 27.841 & 0.049 & 0.136 & 0.340 \\
 SimplE & 0.429 & 11.760 & 0.275 & 0.506 & 0.736 & 0.177  & 31.965 & 0.091 & 0.174 & 0.354 \\
 MINERVA & 0.580 & -- & 0.448 & 0.657 & \bf{0.857}  & 0.174 & -- &  \bf{0.097} & 0.18 & 0.364 \\
 R2D2 & \bf{0.589} & \bf{6.332} & \bf{0.459} & \bf{0.665} & 0.853  & \bf{0.206} & \bf{23.486} & 0.090 & \bf{0.219} & \bf{0.455} 
 
\end{tabular}
}
\caption{The performance on the KG completion task.}
\label{tbl:results}
\end{table*}
Table \ref{tbl:results} displays the results on the completion task for all methods under consideration on FB15k-237 and Hetionet (subsets; see above). R2D2 outperforms all other methods on FB15k-237 with respect to all metrics but Hits@10. However, the performance of MINERVA is almost on par. Moreover, R2D2 outperforms all baselines on Hetionet with respect to the MRR, the mean rank, Hits@3, and Hits@10. While MINERVA exhibits the best performance with respect to Hits@1, R2D2 yields significantly better results with respect to all other metrics.


\begin{table*}
\center
\resizebox{0.85\textwidth}{!}{
\begin{tabular}{ c c | c}
\hline
 \textbf{Query:}
  &  Richard Feynman $\xrightarrow{nationality}$ USA? & Nelson Mandela $\xrightarrow{hasProfession}$ Actor? \\  
  \hline
  
\textbf{Agent 1:} 
 & Richard Feynman $\xrightarrow{livedInLocation}$ Queens &  Nelson Mandela $\xrightarrow{hasFriend}$ Naomi Campbell \\
 & $\wedge$ Queens $\xrightarrow{locatedIn}$ USA & Naomi Campbell  $\xrightarrow{hasDated}$ Leonardo DiCaprio \\
   \hline
\textbf{Agent 2:} & Richard Feynman $\xrightarrow{hasEthnicity}$ Russian people  & Nelson Mandela $\xrightarrow{hasProfession}$ Lawyer \\
 & $\wedge$ Russian people$\xrightarrow{geographicDistribution}$ Republic of Tajikistan  & $\wedge$ Lawyer $\xrightarrow{specializationOf^{-1}}$ Barrister 
\end{tabular}
}
\caption{Two example debates generated by R2D2: While agent 1 argues that the query is true and agent 2 argues that it is false.}
\label{tab:deb_example}
\end{table*}

\paragraph{Survey} To asses whether the arguments are informative for users in an objective setting, we conducted a survey where respondents take the role of the judge making a classification decision based on the agents' arguments. More concretely, we set up an online quiz consisting of ten rounds. Each round is centered around a query (with masked subject) sampled from the test set of FB15k-237 (KG completion). Along with the query statement we present the users six arguments extracted by the agents in randomized order. Based on these arguments the respondents are supposed to judge whether the statement is true or false. In addition, we  asked the respondents to rate their confidence in each round. 

Based on 44 participants (109 invitations were sent) we find that the overall accuracy of the respondents' classifications was 81.8\%. Moreover, based on a majority vote (i.e., classification based on the majority of respondents) nine out of ten questions were classified correctly indicating that humans are approximately on par with the performance of the automated judge. Further, the statement where the majority of respondents was wrong corresponds to the only query that was also misclassified by the judge. In this round the participants were supposed to decide whether a person has the religion Methodism. It is hard to answer this question correctly because the person at hand is Margaret Thatcher who had two different religions over her lifetime: Methodism and the Church of England. The fact that the majority of respondents and the judge agree in all rounds indicates that the judge is aligned with human intuition and that the arguments are informative. Moreover, we  found that when users assigned a high confidence score to their decision ('rather certain' or 'absolutely certain') the overall accuracy of their classification was 89\%. The accuracy dropped to 68.4\% when users assigned a low confidence score ('rather uncertain' or 'absolutely uncertain'). The complete survey along with the a detailed evaluation is reported in the supplementary material.


\section{Discussion and Future Works}
\label{sec:discussion}
We examined the quality of the extracted paths manually and typically found reasonable arguments, but quite often also arguments that do not make intuitive sense. We conjecture that one reason for that is that agents often have difficulties finding meaningful evidence if they are arguing for the false position. Moreover, for many arguments, most of the relevant information is already contained in the first step of the agents and later transitions often contain seemingly irrelevant information. This phenomenon might be due to the fact that the judge ignores later transitions and agents do not receive meaningful rewards. Further, relevant information about the neighborhood of entities can be encoded in the embeddings of  entities. While the judge has access to this information through the training process, it remains hidden to users. For example, when arguing that Nelson Mandela was an actor (see Table \ref{tab:deb_example}) 
the argument of agent 1 requires the user to know that Naomi Campbell and Leonardo DiCaprio are actors (which is encoded in FB15k-237). Then this argument serves as evidence that Nelson Mandela was also an actor since people tend to have friends that share their profession (social homophily). However, without this context information it is not intuitively clear why this is a reasonable argument. 

To examine the interplay between the agents and the judge, we consider a setting where we train R2D2 with two agents, but neglect the arguments of one agent during testing. This should lead to a biased outcome in favor of the agent whose arguments are considered by the judge. We test this setting on FB15k-237 and find that when we consider only the arguments of agent 1, the number of positive predictions increases by 18.8\%. In contrast, when we only consider agent 2, the number positive predictions drops by 70.2\%. This result shows that the debate dynamics are functioning as intended and that the agents learn to extract arguments that the judge considers as evidence for their respective position.

While the results of the survey are encouraging, we plan to develop variants of R2D2 that improve the quality of the arguments and conduct a large scale experimental study that also includes other baselines in a controlled setting. Moreover, we plan to discuss fairness and responsibility considerations. In that regard, \cite{nickel2015review} stress that when applying statistical methods to incomplete KGs  the results are likely to be affected by biases in the data generating process and should be interpreted accordingly. Otherwise, blindly following these predictions can strengthen the bias. While the judge in our method also exploits skews in the data, the arguments can help to identify these biases and potentially exclude problematic arguments from the decision. 


\section{Conclusion}
\label{sec:conclusion}
We proposed R2D2, a new approach for KG reasoning based on a debate game between two opposing reinforcement learning agents. The agents search the KG for arguments that convince a binary classifier of their position. Thereby, they act as sparse, adversarial feature generators. Since the classifier (judge) bases its decision solely on mined arguments, R2D2 is more interpretable than other baseline methods.
Our experiments showed that R2D2 outperforms all baselines in the triple classification setting with respect to all metrics on the benchmark datasets WN18RR and FB15k-237. Moreover, we demonstrated that R2D2 can in principle operate in the KG completion setting. We found that R2D2 has competitive performance compared to all baselines on a subset of FB15k-237 and Hetionet. Furthermore, the results of our survey indicate that the arguments are informative and that the judge is aligned with human intuition.


\section*{Acknowledgement}

The authors acknowledge the support by the German Federal Ministry for Education and Research (BMBF), funding project “MLWin” (grant 01IS18050).
\begin{figure}[h!]
 \begin{center}
    \includegraphics[width=0.18\textwidth]{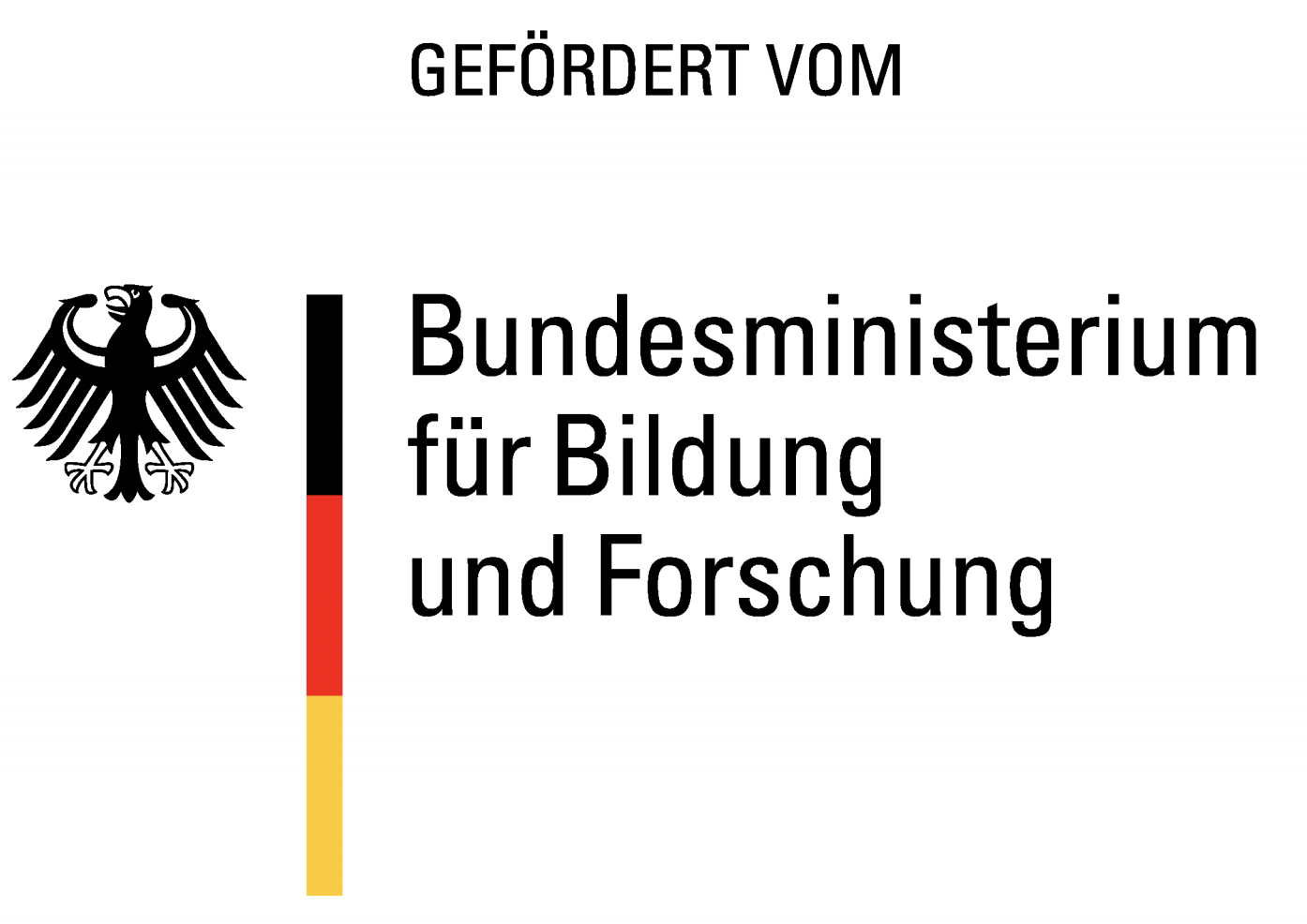}
 \end{center}
\end{figure}

\bibliography{bibliography}
\bibliographystyle{aaai}


\section*{Supplementary material}

# Supplementary Material

## Hyperparameters for R2D2

The number of the latent dimensions $d$ for the embeddings is chosen from the range $\{32, 64, 128\}$. The number of LSTM layers for the agents is chosen among $\{1, 2, 3\}$. The the number of layers in the MLP for the judge is tuned from $\{1, 2, 3, 4, 5\}$. $\beta$ was chosen among $\{0.02, 0.05, 0.1\}$. The length of each argument $T$ was tuned from the range $\{1, 2, 3\}$ and the number of debate rounds $N$ was set to 3. Moreover, the $L_2$-regularization strength $\lambda$ is set to 0.02. Furthermore, the number of rollouts during training is given by 20 and 50 at test time. The loss of the judge and the rewards of the agents were optimized using Adam with learning rate given $10^{-4}$. The best hyperparameter are reported in Table 1.

| Parameter | FB15k-237 | WN18RR | FB15k-237 (subset) | Hetionet |
|---|---|---|---|---|
| Embedding size ($d$) | 64 | 64 | 64 | 32 |
| # stacked LSTM cells (agents) | 2 | 1 | 2 | 2 |
| # layers MLP (judge) | 1 | 1 | 3 | 2 |
| # rounds in a debate ($N$) | 3 | 3 | 3 | 3 |
| Argument/path length (T) | 2 | 2 | 2 | 2 |
| $L_2$-regularization ($\lambda$) | 0.02 | 0.02 | 0.02 | 0.02 |
| Entropy regularization ($\beta$) | 0.02 | 0.02 | 0.1 | 0.1 |
| Learning rate (agents) | $10^{-4}$ | $10^{-4}$ | $10^{-4}$ | $10^{-4}$ |
| Learning rate (judge) | $10^{-4}$ | $10^{-4}$ | $10^{-4}$ | $10^{-4}$ |
| Training rollouts | 20 | 20 | 20 | 20 |
| Test rollouts | 50 | 50 | 100 | 100 |

Table 1: The best hyperparameters for R2D2 found via a cross-validation.

## Hyperparameters for the embedding based baseline methods

**TransE** : The number of the latent dimensions is chosen from the range $\{32, 64, 128, 256\}$. Moreover, the margin $\gamma$ for the loss is tuned from the range $\{0.1, 1, 2\}$. Furthermore, the number of negative triples is fixed to 1. We employed AdaGrad with learning rate 0.01 to fit the model.

| Parameter | FB15k-237 | WN18RR | FB15k-237 (subset) | Hetionet |
|---|---|---|---|---|
| Embedding size | 32 | 32 | 64 | 256 |
| Margin ($\gamma$) | 0.1 | 2 | 1 | 1 |
| Learning rate | 0.01 | 0.01 | 0.01 | 0.01 |
| # negative triples | 1 | 1 | 1 | 1 |

Table 2: The best hyperparameters for TransE found via a cross-validation.

**TransR** : The number of the latent dimensions is chosen from the range $\{64, 128, 256\}$. Moreover, the margin $\gamma$ for the loss is tuned from the range $\{1, 2, 4\}$. Furthermore, the number of negative triples is fixed to 10. We employed AdaGrad with learning rate dependant on the reported value in the original publication.

| Parameter | FB15k-237 | WN18RR | FB15k-237 (subset) | Hetionet |
|---|---|---|---|---|
| Embedding size | 64 | 128 | 128 | 64 |
| Margin ($\gamma$) | 4 | 2 | 2 | 2 |
| Learning rate | 0.001 | 0.1 | 0.01 | 0.01 |
| # negative triples | 10 | 10 | 10 | 10 |

Table 3: The best hyperparameters for TransR found via a cross-validation.

**DistMult** : The number of the latent dimensions is chosen from the range $\{32, 64, 128, 200, 512\}$. Moreover, the $L_2$-regularization strength $\lambda$ is tuned from the range $\{0.001, 0.01, 0.3\}$. Furthermore, the number of negative triples is fixed to 1. We employed AdaGrad with learning rate 1.0 for WN18RR and 0.5 for the other datasets.

| Parameter | FB15k-237 | WN18RR | FB15k-237 (subset) | Hetionet |
|---|---|---|---|---|
| Embedding size | 32 | 32 | 256 | 32 |
| $L_2$-regularization ($\lambda$) | 0.01 | 0.01 | 0.3 | 0.01 |
| Learning rate | 0.5 | 1.0 | 0.5 | 0.5 |
| # negative triples | 10 | 1 | 10 | 10 |

Table 4: The best hyperparameters for DistMult found via a cross-validation.

**ComplEx** : The number of the latent dimensions is chosen from the range $\{32, 64, 128, 256\}$. Moreover, the $L_2$-regularization strength $\lambda$ is tuned from the range $\{0.001, 0.01, 0.3\}$. Furthermore, the number of negative triples is fixed to 10. We employed AdaGrad with learning rate 0.5 to fit the model.

**SimplE** : The number of the latent dimensions is chosen from the range $\{64, 128, 256\}$. Moreover, the $L_2$-regularization strength $\lambda$ is tuned from the range $\{0.01, 0.03, 0.1\}$. Furthermore, the number of negative triples is fixed to 10. We employed AdaGrad with learning rate of either 0.01 or 0.05, depending on the reported learning model in the original publication.

**MINERVA** : The number of the latent dimensions for the embeddings is chosen from the range $\{32, 50, 128\}$. The number of LSTM layers is chosen

| Parameter | FB15k-237 | WN18RR | FB15k-237 (subset) | Hetionet |
|---|---|---|---|---|
| Embedding size | 32 | 64 | 128 | 32 |
| $L_2$-regularization ($\lambda$) | 0.01 | 0.3 | 0.3 | 0.01 |
| Learning rate | 0.5 | 0.5 | 0.5 | 0.5 |
| # negative triples | 10 | 10 | 10 | 10 |

Table 5: The best hyperparameters for ComplEx found via a cross-validation.

| Parameter | FB15k-237 | WN18RR | FB15k-237 (subset) | Hetionet |
|---|---|---|---|---|
| Embedding size | 256 | 128 | 256 | 256 |
| $L_2$-regularization ($\lambda$) | 0.01 | 0.03 | 0.01 | 0.3 |
| Learning rate | 0.05 | 0.05 | 0.05 | 0.05 |
| # negative triples | 10 | 10 | 10 | 10 |

Table 6: The best hyperparameters for SimplE found via a cross-validation.

among $\{1, 2, 3\}$. The entropy regularizer $\beta$ was chosen among $\{0.01, 0.1\}$ and the $L_2$-regularization strength $\lambda$ is set to 0.02. The maximal path length was tuned from the range $\{2, 3\}$. For Hetionet, we also considered path lengths 4 and 5. Furthermore, the number of rollouts during training is given by 20 and a beam width of 100 was employed during testing. The loss of the judge and the rewards of the agents were optimized using Adam with learning rate given 0.001. The best hyperparameter are reported in Table 7.

| Parameter | FB15k-237 (subset) | Hetionet |
|---|---|---|
| Embedding size ($d$) | 50 | 32 |
| # LSTM cells | 2 | 2 |
| Path length (T) | 3 | 5 |
| $L_2$-regularization ($\lambda$) | 0.02 | 0.02 |
| Entropy regularization ($\beta$) | 0.1 | 0.01 |
| Learning rate | 0.001 | 0.001 |
| Training rollouts | 20 | 20 |
| Beam width | 100 | 100 |

Table 7: The best hyperparameters for MINERVA found via a cross-validation.

## Survey

Below we included the online survey (see Section 4 of the paper) followed by an evaluation of every round based on 45 participants (109 invitations were sent). In order not to bias the respondents, we did not inform them about the type of project that we have been working on. Further, all statements and facts are generated and permuted at random and no post-processing was applied. Automatically filtering arguments, ranking them according to their score, and displaying the arguments in a structured way (e.g., matching them to the agents) leads to a clearer outcome. However, in order to create an objective setting, we did not consider any such steps.

# AI Quiz

We are a group of researchers and engineers working on an interactive question answering tool. First, we want to thank you for supporting our work! We realize how precious your time is. That's why we made sure this quiz will take at most 10 minutes.

* Required

## Instruction
You will participate in a quiz consisting of ten rounds. Each round is centered around a statement about an anonymized celebrity (think of a random person from Wikipedia). The statement can be either true or false.

Along with the statement we show you six facts. Based on these facts your task is to judge whether the statement is true or false. While some facts may be informative, others contain useless information. Don't worry if you feel that you cannot make an informed decision: Guessing is part of this game!

## Additional Remarks
We also ask you to rate your confidence of each evaluation.

Please don't look for external information (e.g., Google, Wikipedia) or talk to other respondents about the quiz unless you need vocabulary clarifications or basic definitions

Note that all statements and facts are accumulated over time. Thus, seemingly conflicting statements and facts can be true (e.g., people can in principle have multiple nationalities, religions, etc.).

You can leave comments for us at the end of this quiz.

## Statement: Person 1 has female gender.

Fact 1: Person 1 was nominated for an award for the Best Female Lead. Joan Allen was also nominated for an award for the Best Female Lead.

Fact 2: Person 1 was nominated for the Primetime Emmy Award for Outstanding Lead Actress in a Comedy Series. Malcom is the Middle was also nominated for the Primetime Emmy Award for Outstanding Lead Actress in a Comedy Series.

Fact 3: Person 1 was nominated for an award together with Teri Hatcher.

Fact 4: Person 1 was nominated for an award together with Teri Hatcher. Teri Hatcher dated Ryan Seacrest.

Fact 5: Person 1 performed in the film Magnolia. Henry Gibson also performed in the film Magnolia.

Fact 6: Person 1 was nominated for an Oscar for the Best Actress. Sandra Bullock was also nominated for an Oscar for the Best Actress.

1. **Evaluation 1 ***
    *Mark only one oval.*

    ( ) The statement is true.

    ( ) The statement is false.

2. **How confident are you about evaluation 1?** *

   *Mark only one oval.*

   ( ) I am absolutely certain.

   ( ) I am rather certain.

   ( ) I am rather uncertain.

   ( ) I am absolutely uncertain (random guess).

## Statement: Person 2 has the profession engineer.

Fact 1: Person 2 was nominated for the Outstanding Comedy Series Award. Murphy Brown was also nominated for the Outstanding Comedy Series Award.

Fact 2: Person 2 was nominated for an award together with Sam Means. Sam Means has the profession graphic humorist.

Fact 3: Person 2 has the profession writer. Rutger Hauer also has the profession writer.

Fact 4: Person 2 has the nationality United States of America. The film A Civil Action has the country of origin United States of America.

Fact 5: Person 2 was nominated for an award together with Ron Weiner. Ron Weiner was also nominated for an award together with David Miner.

Fact 6: Person 2 was nominated for an award together with Alec Baldwin. Alec Baldwin graduated from the Lee Strasberg Theatre and Film Institute.

3. **Evaluation 2** *

   *Mark only one oval.*

   ( ) The statement is true.

   ( ) The statement is false.

4. **How confident are you about evaluation 2?** *

   *Mark only one oval.*

   ( ) I am absolutely certain.

   ( ) I am rather certain.

   ( ) I am rather uncertain.

   ( ) I am absolutely uncertain (random guess).

## Statement: Person 3 has Jewish ethnicity.

Fact 1: Person 3 dated Keira Knightley. Keira Knightley has the profession model.

Fact 2: Person 3 was nominated for an award for Cadillac Records. Cadillac Records had a film crew role for visual effects.

Fact 3: Person 3 lived in Queens. Queens is located in the United States of America

Fact 4: Person 3 was nominated for an award together with Michael Sheen. Michael Sheen has the nationality Wales.

Fact 5: Person 3 graduated from Queens College, City University of New York. Ron Jeremy also graduated from Queens College, City University of New York.

Fact 6: Person 3 won an award for The Pianist. The Pianist has the release Region United Kingdom.

5. **Evaluation 3** *

   *Mark only one oval.*

   ( ) The statement is true.

   ( ) The statement is false.

6. **How confident are you about evaluation 3?** *

   *Mark only one oval.*

   ( ) I am absolutely certain.

   ( ) I am rather certain.

   ( ) I am rather uncertain.

   ( ) I am absolutely uncertain (random guess).

## Statement: Person 4 has nationality Austrian.

Fact 1: Person 4 was the director of the film From Here to Eternity.

Fact 2: Person 4 is the director of the film From Here to Eternity. Ernest Borgnine performed in the film From Here to Eternity.

Fact 3: Person 4 produced the film A Man for All Seasons. A Man for All Seasons is a netflix title in the United Kingdom.

Fact 4: Person 4 produced the film A Man for All Seasons. A Man for All Seasons has the country of origin United Kingdom.

Fact 5: Person 4 was born in Vienna. Arnold Schönberg lived in Vienna.

Fact 6: Person 4 lived in Vienna. Arnold Schönberg also lived in Vienna.

7. **Evaluation 4** *

   *Mark only one oval.*

   ( ) The statement is true.

   ( ) The statement is false.

8. **How confident are you about evaluation 4?** *

   *Mark only one oval.*

   ( ) I am absolutely certain.

   ( ) I am rather certain.

   ( ) I am rather uncertain.

   ( ) I am absolutely uncertain (random guess).

## Statement: Person 5 has religion Calvinism.

Remark: Calvinism is a branch of Protestant Christianity.

Fact 1: Person 5 is married. Charo Santos-Concio is also married.

Fact 2: Person 5 performed in the film Jerry Maguire. James L. Brooks also performed in the film Jerry Maguire.

Fact 3: Person 5 performed in the film Gettysburg. The film Gettysburg has the country of origin United States of America.

Fact 4: Person 5 has religion Catholicism. David Wenham also has religion Catholicism.

Fact 5: Person 5 performed in the film American Splendor. Fine Line Features was nominated for an award for the film American Splendor.

Fact 6: Person 5 performed in the film Little Women. Winona Ryder was nominated for an award for the film Little Women.

9. **Evaluation 5** *

   *Mark only one oval.*

   ( ) The statement is true.

   ( ) The statement is false.

10. **How confident are you about evaluation 5?** *

    *Mark only one oval.*

    ( ) I am absolutely certain.

    ( ) I am rather certain.

    ( ) I am rather uncertain.

    ( ) I am absolutely uncertain (random guess).

## Statement: Person 6 has religion Methodism.

Remark: Methodism is a group of Protestant Christianity.

Fact 1: Person 6 held the government position Prime Minister. Vietnam also has the government position Prime Minister.

Fact 2: Person 6 has the profession scientist. Stephen Hawkins is also a scientist.

Fact 3: Person 6 was influenced by Friedrich Hayek.

Fact 4: Person 6 has the religion Church of England.

Fact 5: Person 6 has ethnicity caucasian. The ethnicity caucasian is a risk factor for testicular cancer.

Fact 6: Person 6 held a government position in the United Kingdom. The United Kingdom is a release region of the film X-Men: First Class.

11. **Evaluation 6** *

    *Mark only one oval.*

    ( ) The statement is true.

    ( ) The statement is false.

12. **How confident are you about evaluation 6?** *

    *Mark only one oval.*

    ( ) I am absolutely certain.

    ( ) I am rather certain.

    ( ) I am rather uncertain.

    ( ) I am absolutely uncertain (random guess).

## Statement: Person 7 has nationality Austrian Empire.

Fact 1: Person 7 has the profession hairdresser. The film The World Is Not Enough has the crew role hairdresser.

Fact 2: Person 7 is burried at the Golders Green Crematorium.

Fact 3: Person 7 died in Sunderland.

Fact 4: Person 7 has nationality England. Timothy Spall also has nationality England.

Fact 5: Person 7 died in Sunderland. Sunderland is located in the United Kingdom

Fact 6: Person 7 has nationality South Africa. South Africa is a release region for the film Men in Black 3.

13. **Evaluation 7** *

    *Mark only one oval.*

    ( ) The statement is true.

    ( ) The statement is false.

14. **How confident are you about evaluation 7?** *

    *Mark only one oval.*

    ( ) I am absolutely certain.

    ( ) I am rather certain.

    ( ) I am rather uncertain.

    ( ) I am absolutely uncertain (random guess).

## Statement: Person 8 has ethnicity Armenian.

Fact 1: Person 8 is friend of Michael Jackson. Michael Jackson was nominated for a Grammy Award for Best Male R&B Vocal Performance.

Fact 2: Person 8 graduated from the Professional Children's School. John Spencer also graduated from the Professional Children's School.

Fact 3: Person 8 won the MTV Movie Award for Best Kiss. Annette Bening also won the MTV Movie Award for Best Kiss.

Fact 4: Person 8 was nominated for an award for the Worst Actor. The film Jack and Jill was also nominated in the category Worst Actor.

Fact 5: Person 8 is friend of Michael Jackson. Michael Jackson has the profession songwriter.

Fact 6: Person 8 performed in the film Home Alone. The film Home Alone has the release region Singapur.

15. **Evaluation 8** *

    *Mark only one oval.*

    ( ) The statement is true.

    ( ) The statement is false.

16. **How confident are you about evaluation 8?** *

    *Mark only one oval.*

    ( ) I am absolutely certain.

    ( ) I am rather certain.

    ( ) I am rather uncertain.

    ( ) I am absolutely uncertain (random guess).

## Statement: Person 9 has female gender.

Fact 1: Person 9 is an artist of the genre Pop. Daddy Yankee is also an artist of the genre Pop.

Fact 2: Person 9 is an artist of the genre Country Pop. The Nitty Gritty Dirt Band is an artist group of the genre Country Pop.

Fact 3: Person 9 is an artist with the record label Atlantic Records. John Lee Hooker is also an artist with the record label Atlantic Records.

Fact 4: Person 9 won an award for the Best Male Country Vocal Performance. Best Male Country Vocal Performance was an award category at the 52nd Annual Grammy Awards.

Fact 5: Person 9 was nominated for a Grammy Award for the Best Male Pop Vocal Performance . Barry Manilow won a Grammy Award for the Best Male Pop Vocal Performance.

Fact 6: Person 9 has the profession composer. Miles Davis also has the profession composer.

17. **Evaluation 9** *

    *Mark only one oval.*

    ◯ The statement is true.

    ◯ The statement is false.

18. **How confident are you about evaluation 9?** *

    *Mark only one oval.*

    ◯ I am absolutely certain.

    ◯ I am rather certain.

    ◯ I am rather uncertain.

    ◯ I am absolutely uncertain (random guess).

## Statement: Person 10 has the profession author.

Fact 1: Person 10 was nominated for an award for The Goodbye Girl. The Goodbye Girl was also nominated for the British Academy Film Award for the best actress.

Fact 2: Person 10 graduated from DeWitt Clinton High School. The headquarter of DeWitt Clinton High School is located in the Bronx.

Fact 3: Person 10 has the profession songwriter. Lorenz Hart has also the profession songwriter.

Fact 4: Person 10 was nominated for a Golden Globe for the Best Screenplay. A Man for All Seasons won a Golden Globe for the Best Screenplay.

Fact 5: Person 10 was nominated for the Oscar for the Best Original Screenplay. Lost in Translation won the Oscar for the Best Original Screenplay.

Fact 6: Person 10 has the profession dramaturg. Nikolái Gógol is also a dramaturg.

19. **Evaluation 10** *

    *Mark only one oval.*

    ◯ The statement is true.

    ◯ The statement is false.

20. **How confident are you about evaluation 10?** *

    *Mark only one oval.*

    ◯ I am absolutely certain.

    ◯ I am rather certain.

    ◯ I am rather uncertain.

    ◯ I am absolutely uncertain (random guess).

## Thank you for the support!

If you want to be informed about the results of this survey and receive more details about our research you can enter your email adress below.

Further, you can also leave a comment on this quiz (e.g., suggestions, flaws, etc.).

| Round | Truth Value |
|---|---|
| 1 | True |
| 2 | False |
| 3 | True |
| 4 | True |
| 5 | False |
| 6 | True |
| 7 | False |
| 8 | False |
| 9 | False |
| 10 | True |

Table 8: The ground truths of the statements in each round.

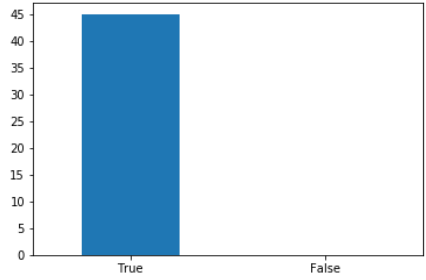

Figure 1: Evaluation 1

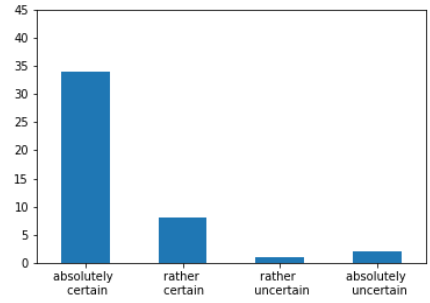

Figure 2: How confident are you about evaluation 1?

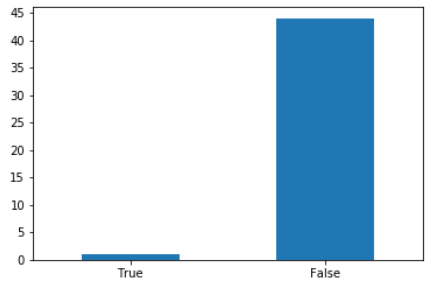

Figure 3: Evaluation 2

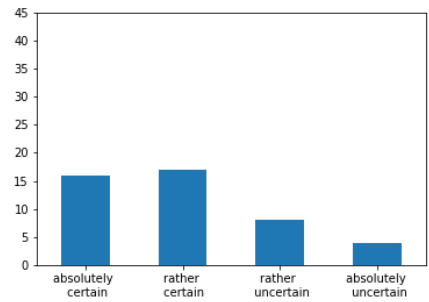

Figure 4: How confident are you about evaluation 2?

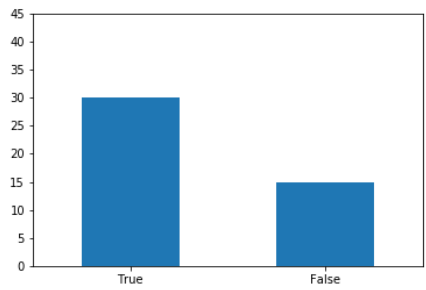

Figure 5: Evaluation 3

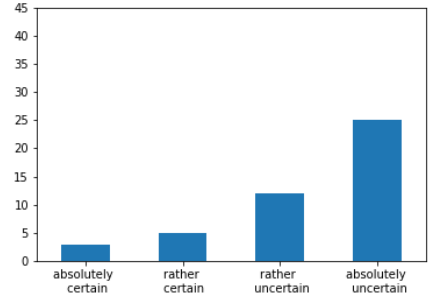

Figure 6: How confident are you about evaluation 3?

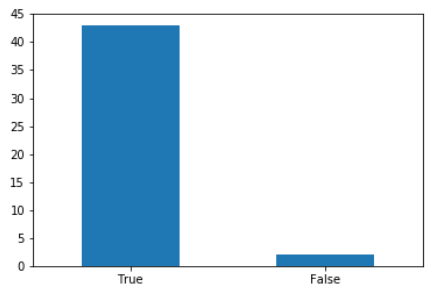

Figure 7: Evaluation 4

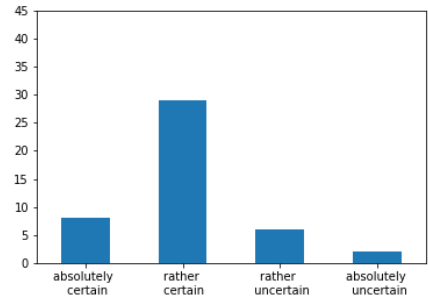

Figure 8: How confident are you about evaluation 4?

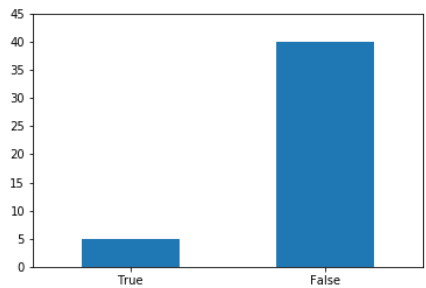

Figure 9: Evaluation 5

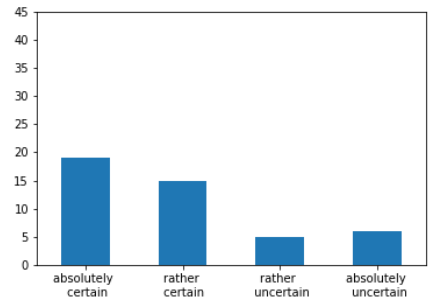

Figure 10: How confident are you about evaluation 5?

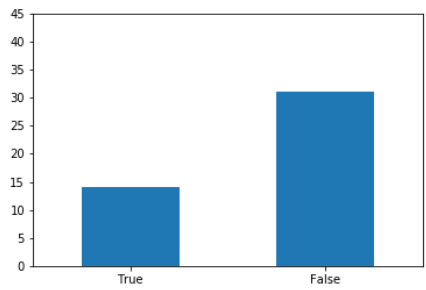

Figure 11: Evaluation 6

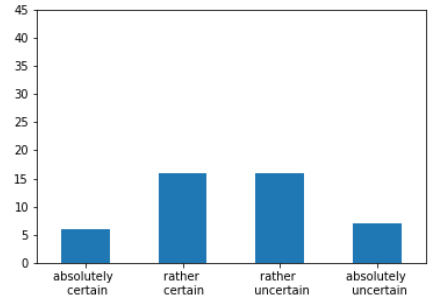

Figure 12: How confident are you about evaluation 6?

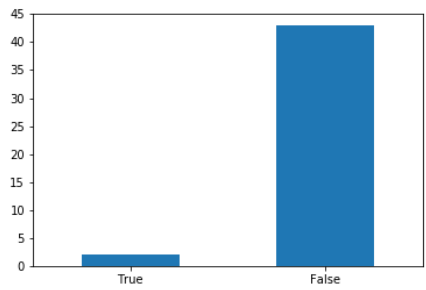

Figure 13: Evaluation 7

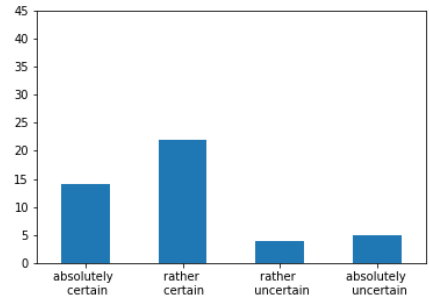

Figure 14: How confident are you about evaluation 7?

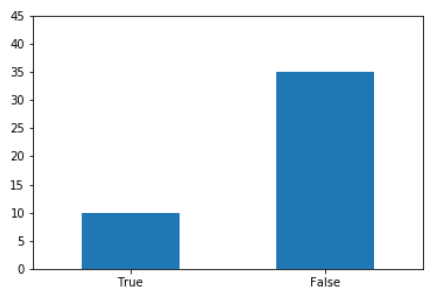

Figure 15: Evaluation 8

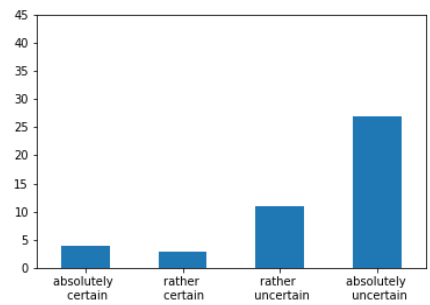

Figure 16: How confident are you about evaluation 8?

and inference in knowledge bases. In *International Conference on Learning Representations*.

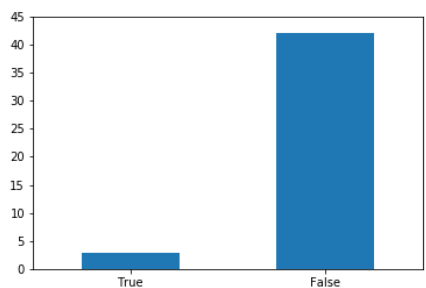

Figure 17: Evaluation 9

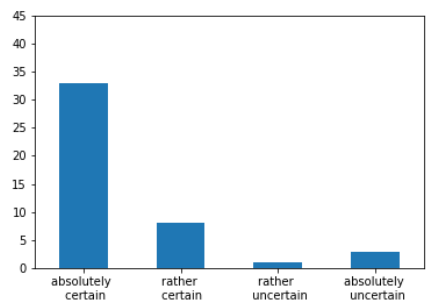

Figure 18: How confident are you about evaluation 9?

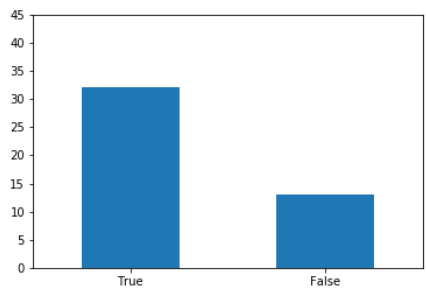

Figure 19: Evaluation 10

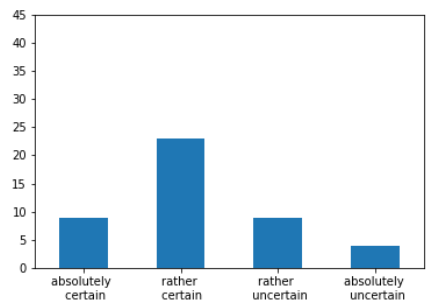

Figure 20: How confident are you about evaluation 10?

14


\end{document}